\documentclass[letterpaper, 10pt, conference]{ieeeconf}  

\IEEEoverridecommandlockouts                              

\usepackage{amsmath} 
\usepackage{amssymb}  
\usepackage{cite}
\usepackage{graphicx}
\usepackage{subcaption}
\usepackage{tikz}
\usetikzlibrary{shapes.geometric, arrows, positioning, calc}

\tikzstyle{blocktheirs} = [rectangle, rounded corners, text centered, draw=black, fill=blue!30]
\tikzstyle{blockours} = [rectangle, rounded corners, text centered, draw=black, fill=green!30]
\tikzstyle{graph} = [rectangle, rounded corners, text centered, draw=black, fill=orange!30]
\tikzstyle{arrow} = [thick,->,>=stealth]

\usepackage{pgfplots}
\pgfplotsset{compat=1.3}

\usepackage{algorithm}
\usepackage{algorithmic}
\usepackage{orcidlink}
\usepackage{multirow}
\usepackage{colortbl}
\usepackage{hhline}
\usepackage[T1]{fontenc}
\usepackage{inconsolata}

\usepackage{siunitx}

\usepgfplotslibrary{groupplots}
\usepackage{csquotes}

\usepackage{mathrsfs}
\usepackage[nomain,acronym,toc,indexonlyfirst=true]{glossaries}
\makeatletter

\newglossarystyle{alttreeMRM}{%
  \setglossarystyle{alttree}%
  \renewenvironment{theglossary}%
    {\def\@gls@prevlevel{-1}%
     \mbox{}\par}%
    {\par}%
  \renewcommand*{\glsgroupheading}[1]{}%
}

\newglossarystyle{MRMSymbols}{
  \setglossarystyle{alttreeMRM}%
  
  \renewenvironment{theglossary}%
    {\def\@gls@prevlevel{-1}%
     \setlength{\parskip}{.5ex}%
     \mbox{}\par}%
    {\par}%
  \renewcommand{\glsgroupheading}[1]{\par
    \def\@gls@prevlevel{-1}%
    \hangindent0pt\relax
    \parindent0pt\relax
    \textbf{\glsgetgrouptitle{##1}}\par\vspace{.5ex}}%
}

\newglossarystyle{MRMAbbreviations}{%
  \setglossarystyle{alttreeMRM}%

    {\def\@gls@prevlevel{-1}%
     \setlength{\parskip}{.5ex}%
     \setlength{\parfillskip}{\parindent}%
     \mbox{}\par}%
    {\par}%
}

\makeatother

\newacronym{AOI}{AOI}{actor of interest}
\newacronym{CNN}{CNN}{convolutional neural network}
\newacronym{HSV}{HSV}{hue saturation value}
\newacronym{TCN}{TCN}{temporal convolutional network}
\newacronym{L2A}{L2A}{lane-to-actor}
\newacronym{A2L}{A2L}{actor-to-lane}
\newacronym{L2L}{L2L}{lane-to-lane}
\newacronym{A2A}{A2A}{actor-to-actor}
\newacronym{ADE}{ADE}{average displacement error}
\newacronym{NLP}{NLP}{neural language processing}
\newacronym{MHA}{MHA}{multi-head attention}
\newacronym{GCN}{GCN}{graph convolutional network}
\newacronym{GNN}{GNN}{graph neural network}
\newacronym{GAT}{GAT}{graph attention network}
\newacronym{MLP}{MLP}{multi-layer perceptron}
\newacronym{PDF}{PDF}{probability density function}
\newacronym{NLL}{NLL}{negative log-likelihood}

\usepackage{mathtools}

\newcommand{\normalconc}{\mathop{|\mkern-2mu|}}

\newcommand{\bigconcchoose}[1]{\def\bigconcsize{}%
  \ifx#1\displaystyle
    \let\bigconcsize\Big
  \else
    \ifx#1\textstyle
      \let\bigconcsize\big
    \fi
  \fi#1}

\newglossary[slg]{symbols}{sym}{sbl}{\iflanguage{english}%
{List of Symbols}{Symbolverzeichnis}}

\newglossaryentry{Real}{name=\ensuremath{\mathbb{R}},description={Space of all real numbers},sort=gr, type=symbols}
\newglossaryentry{Ds}{name=\ensuremath{d_s},description={Dimension of a state variable},sort=gds, type=symbols}
\newglossaryentry{Dfn}{name=\ensuremath{d_f},description={Dimension of a node feature vector},sort=gdf, type=symbols}
\newglossaryentry{De}{name=\ensuremath{d_e},description={Dimension of an edge feature vector},sort=gdf, type=symbols}
\newglossaryentry{I}{name=\ensuremath{I_N},description={Identity matrix of shape $N \times N$},sort=gI, type=symbols}
\newglossaryentry{Deg}{name=\ensuremath{D},description={Degree matrix},sort=gI, type=symbols}
\newglossaryentry{state}{name=\ensuremath{S},description={State of an actor},sort=gstate, type=symbols}
\newglossaryentry{stateit}{name=\ensuremath{S_{i,t}},description={State of actor $i$ at time step $t$},sort=gstateit, type=symbols}
\newglossaryentry{predstateit}{name=\ensuremath{\hat{S}_{i,t}},description={predicted State of actor $i$ at time step $t$},sort=gstateit, type=symbols}
\newglossaryentry{statesall}{name=\ensuremath{S_{\cdot,\cdot}},description={Any state of any actor at any time step},sort=gstatesall, type=symbols}
\newglossaryentry{numstates}{name=\ensuremath{Z},description={Number of state variables contained in an actor's state},sort=gnumstates, type=symbols}
\newglossaryentry{thisti}{name=\ensuremath{T_i},description={Number of past states available for actor $i$},sort=gti, type=symbols}
\newglossaryentry{trajihist}{name=\ensuremath{\tau_{i,\text{hist}}},description={Set of past states of actor $i$ (including the current state)},sort=gtrajihist,type=symbols}
\newglossaryentry{trajigt}{name=\ensuremath{\tau_{i,\text{gt}}},description={Set of ground truth states of actor $i$ (not including the current state)},sort=gtrajigt,type=symbols}
\newglossaryentry{trajgt}{name=\ensuremath{\tau_{\text{gt}}},description={A set of ground truth states for an unspecified actor},sort=gtrajigt,type=symbols}
\newglossaryentry{trajifut}{name=\ensuremath{\tau_{i,\text{fut}}},description={Set of future states of actor $i$ (not including the current state)},sort=gtrajifut,type=symbols}
\newglossaryentry{pathifut}{name=\ensuremath{\mathcal{P}_{i,\text{fut}}},description={List of future path points of actor $i$},sort=gpathifut,type=symbols}
\newglossaryentry{trajpredm}{name=\ensuremath{\hat{\tau}_{i,m}},description={Set of predicted future states of actor $i$ (not including the current state) of predicted mode $m$},sort=gtau, type=symbols}
\newglossaryentry{alltrajihist}{name=\ensuremath{\tau_{\cdot,\text{hist}}},description={Set of past states of all actors (including the current states)},sort=gtrajihistall,type=symbols}
\newglossaryentry{alltrajigt}{name=\ensuremath{\tau_{\cdot,\text{gt}}},description={Set of ground truth states of all actors (not including the current states)},sort=gtrajigtall,type=symbols}
\newglossaryentry{alltrajifut}{name=\ensuremath{\tau_{\cdot,\text{fut}}},description={Set of future states of all actors (not including the current states)},sort=gtrajifutall,type=symbols}
\newglossaryentry{setall}{name=\ensuremath{\mathcal{O}_\text{actors}},description={Set of indices of all perceived actors},sort=goactors,type=symbols}
\newglossaryentry{setpred}{name=\ensuremath{\mathcal{O}_\text{pred}},description={Set of indices of those actors which shall be predicted},sort=gopred,type=symbols}
\newglossaryentry{setfut}{name=\ensuremath{\mathcal{O}_\text{fut}},description={Set of indices of those actors for which cooperative information about their future is available},sort=gofut,type=symbols}
\newglossaryentry{setother}{name=\ensuremath{\mathcal{O}_\text{other}},description={Set of indices of actors which do not have to be predicted},sort=goother,type=symbols}
\newglossaryentry{graph}{name=\ensuremath{\mathcal{G}},description={A graph},sort=gg,type=symbols}
\newglossaryentry{graphact}{name=\ensuremath{\mathcal{G}_\text{act}},description={Graph of actor nodes},sort=ggact,type=symbols}
\newglossaryentry{graphl}{name=\ensuremath{\mathcal{G}_\text{lane}},description={Graph of lane nodes},sort=gglane,type=symbols}
\newglossaryentry{graphtraj}{name=\ensuremath{\mathcal{G}_\text{traj}},description={Graph of future nodes},sort=ggfut,type=symbols}
\newglossaryentry{graphpath}{name=\ensuremath{\mathcal{G}_\text{path}},description={Graph of path nodes},sort=ggpaths,type=symbols}
\newglossaryentry{conns}{name=\ensuremath{\mathcal{C}},description={Set of possible connection types},sort=gconn,type=symbols}
\newglossaryentry{threshconna2l}{name=\ensuremath{\theta_\text{act}},description={Threshold (a euclidean distance) that defines if two nodes are connected or not (for the A2L module)},sort=gtact,type=symbols}
\newglossaryentry{threshconnact}{name=\ensuremath{\theta_\text{act}},description={Threshold (a euclidean distance) that defines if two nodes in the actor graph are connected or not},sort=gtact,type=symbols}
\newglossaryentry{numobjs0}{name=\ensuremath{N_{O,0}},description={Number of objects at the current time step},sort=gnot,type=symbols}
\newglossaryentry{numobjst}{name=\ensuremath{N_{O,t}},description={Number of objects at time step $t$},sort=gnot,type=symbols}
\newglossaryentry{x}{name=\ensuremath{x},description={x-position of an actor},sort=gx, type=symbols}
\newglossaryentry{y}{name=\ensuremath{y},description={y-position of an actor},sort=gy, type=symbols}
\newglossaryentry{w}{name=\ensuremath{w},description={width of an actor},sort=gw, type=symbols}
\newglossaryentry{length}{name=\ensuremath{l},description={length of an actor},sort=gl, type=symbols}
\newglossaryentry{v}{name=\ensuremath{v},description={velocity of an actor},sort=gv, type=symbols}
\newglossaryentry{yaw}{name=\ensuremath{\varphi},description={orientation of an actor},sort=gphi, type=symbols}
\newglossaryentry{yrt}{name=\ensuremath{\dot{\varphi}},description={yaw rate of an actor},sort=gphidot, type=symbols}
\newglossaryentry{H}{name=\ensuremath{H},description={Prediction horizon (number of time steps that are to be predicted)},sort=gh, type=symbols}
\newglossaryentry{T}{name=\ensuremath{T},description={History length (number of time steps for which historic data is available)},sort=gh, type=symbols}
\newglossaryentry{traj}{name=\ensuremath{\tau},description={Trajectory (set of states)},sort=gtau, type=symbols}
\newglossaryentry{map}{name=\ensuremath{\mathcal{M}},description={Map},sort=gmap, type=symbols}
\newglossaryentry{M}{name=\ensuremath{M},description={number of predicted modes},sort=gm, type=symbols}
\newglossaryentry{pmodei}{name=\ensuremath{\hat{p}_{i,m}},description={predicted probability of mode $m$ for predictions of actor $i$},sort=gpim, type=symbols}
\newglossaryentry{pdftrajim}{name=\ensuremath{\hat{p}_{i}(\hat{\tau}_{i,m})},description={probability density function over predicted trajectory mode $m$ of actor $i$},sort=gpdfim, type=symbols}
\newglossaryentry{imgW}{name=\ensuremath{W},description={width of an image},sort=gwimg, type=symbols}
\newglossaryentry{imgH}{name=\ensuremath{H},description={height of an image},sort=ghimg, type=symbols}
\newglossaryentry{imgC}{name=\ensuremath{C},description={number of channels of an image or a 3D tensor},sort=gcimg, type=symbols}
\newglossaryentry{Fd}{name=\ensuremath{D},description={Number of features produced by an encoder that can be used for a trajectory decoder.},sort=gcfd, type=symbols}
\newglossaryentry{Fnode}{name=\ensuremath{F},description={features for a node $n$ in a graph},sort=gf, type=symbols}
\newglossaryentry{Fnoden}{name=\ensuremath{F_n},description={features for a node $n$ in a graph},sort=gfn, type=symbols}
\newglossaryentry{Fnodemat}{name=\ensuremath{\mathbf{F}},description={all node features of a graph, represented as a matrix},sort=gf, type=symbols}
\newglossaryentry{Fedge}{name=\ensuremath{\epsilon},description={features for an edge $e$ in a graph},sort=gee, type=symbols}
\newglossaryentry{Fedgee}{name=\ensuremath{\epsilon_e},description={features for an edge $e$ in a graph},sort=gee, type=symbols}
\newglossaryentry{Fi}{name=\ensuremath{F_i},description={features for actor $i$},sort=gfi, type=symbols}
\newglossaryentry{Ftcni}{name=\ensuremath{F_{\text{TCN},i}},description={TCN features for actor $i$},sort=gftcni, type=symbols}
\newglossaryentry{Fmapglob}{name=\ensuremath{F_{\text{map},\text{global}}},description={Global map features},sort=gfmapglob, type=symbols}
\newglossaryentry{Fmaploc}{name=\ensuremath{F_{\text{map},\text{local}}},description={Local map features},sort=gfmaploc, type=symbols}
\newglossaryentry{Fmaploci}{name=\ensuremath{F_{\text{map},\text{local},i}},description={Local map features for actor $i$},sort=gfmaploci, type=symbols}
\newglossaryentry{Nd}{name=\ensuremath{\mathcal{N}},description={Set of nodes of a graph},sort=gnd, type=symbols}
\newglossaryentry{NumNd}{name=\ensuremath{N},description={Number of nodes in a graph},sort=gnumnd, type=symbols}
\newglossaryentry{Ndl}{name=\ensuremath{\mathcal{N}_l},description={Lane graph node},sort=gndl, type=symbols}
\newglossaryentry{NumNdl}{name=\ensuremath{N_l},description={Number of lane nodes},sort=gnumndl, type=symbols}
\newglossaryentry{Adj}{name=\ensuremath{\mathcal{A}},description={Adjacency matrix of a graph},sort=gadj, type=symbols}
\newglossaryentry{Adjn1n2}{name=\ensuremath{\mathcal{A}_{n_1,n_2}},description={Adjacency matrix entry for a connection between nodes $n_1$ and $n_2$},sort=gadjn1n2, type=symbols}
\newglossaryentry{Adjpre}{name=\ensuremath{\mathcal{A}_\text{pre}},description={Adjacency matrix for \emph{predecessor} relationship},sort=gadj, type=symbols}
\newglossaryentry{Adjsuc}{name=\ensuremath{\mathcal{A}_\text{suc}},description={Adjacency matrix for \emph{successor} relationship},sort=gadj, type=symbols}
\newglossaryentry{Adjleft}{name=\ensuremath{\mathcal{A}_\text{left}},description={Adjacency matrix for \emph{left} relationship},sort=gadj, type=symbols}
\newglossaryentry{Adjright}{name=\ensuremath{\mathcal{A}_\text{right}},description={Adjacency matrix for \emph{right} relationship},sort=gadj, type=symbols}
\newglossaryentry{Edges}{name=\ensuremath{\mathcal{E}},description={Set of edges},sort=gedgesl, type=symbols}
\newglossaryentry{NumE}{name=\ensuremath{E},description={Number of edges in a graph},sort=ge, type=symbols}
\newglossaryentry{EdgesL}{name=\ensuremath{\mathcal{E}_l},description={Set of edges},sort=gedgesl, type=symbols}
\newglossaryentry{EdgesA}{name=\ensuremath{\mathcal{E}_a},description={Set of edges in the actor graph},sort=gedgesa, type=symbols}
\newglossaryentry{EdgesL2A}{name=\ensuremath{\mathcal{E}_\text{L2A}},description={Set of edges for the L2A fusion},sort=gedgesl2a, type=symbols}
\newglossaryentry{EdgesA2L}{name=\ensuremath{\mathcal{E}_\text{A2L}},description={Set of edges for the A2L fusion},sort=gedgesa2l, type=symbols}
\newglossaryentry{EdgesA2A}{name=\ensuremath{\mathcal{E}_\text{A2A}},description={Set of edges for the A2A fusion},sort=gedgesa2a, type=symbols}
\newglossaryentry{Nda}{name=\ensuremath{\mathcal{N}_a},description={Actor graph nodes},sort=gnda, type=symbols}
\newglossaryentry{thrasso}{name=\ensuremath{\delta_\text{assoc}},description={Association threshold distance},sort=gnda, type=symbols}
\newglossaryentry{Ri}{name=\ensuremath{R_i},description={Relative location embedding},sort=gri, type=symbols}
\newglossaryentry{raster}{name=\ensuremath{\delta},description={Rasterization, measured in meters per pixel},sort=gd, type=symbols}
\newglossaryentry{veck}{name=\ensuremath{k},description={key vector (in the context of attention)},sort=gk, type=symbols}
\newglossaryentry{vecv}{name=\ensuremath{v},description={value vector (in the context of attention)},sort=gv, type=symbols}
\newglossaryentry{vecq}{name=\ensuremath{q},description={query vector (in the context of attention)},sort=gq, type=symbols}
\newglossaryentry{K}{name=\ensuremath{K},description={Matrix of keys (in the context of attention)},sort=gk, type=symbols}
\newglossaryentry{V}{name=\ensuremath{V},description={Matrix of values (in the context of attention)},sort=gv, type=symbols}
\newglossaryentry{Q}{name=\ensuremath{Q},description={Matrix of queries (in the context of attention)},sort=gq, type=symbols}
\newglossaryentry{dk}{name=\ensuremath{d_k},description={Dimension of a feature vector (in the context of attention)},sort=gdk, type=symbols}
\newglossaryentry{attw}{name=\ensuremath{\alpha_\text{att}},description={Vector of weights calculated by attention},sort=gaatt, type=symbols}
\newglossaryentry{attwm}{name=\ensuremath{A_\text{att}},description={Matrix of weights calculated by attention},sort=gaatt, type=symbols}
\newglossaryentry{attres}{name=\ensuremath{\alpha},description={Attention result (sum of values, weighted by attention weights)},sort=gaatt, type=symbols}
\newglossaryentry{Nheads}{name=\ensuremath{N_h},description={Number of attention heads in multi-head attention},sort=gnh, type=symbols}
\newglossaryentry{deltanu}{name=\ensuremath{\Delta_{n,u}},description={Feature vector that is computed based on the difference in the ordinates of the locations of graph nodes $n$ and $u$},sort=gdnu, type=symbols}
\newglossaryentry{loc}{name=\ensuremath{\mathbf{l}},description={2D vector that contains the (euclidean) location of a node},sort=gloc, type=symbols}

\newglossaryentry{i}{name=\ensuremath{i},description={actor index $i$},sort=si, type=symbols}
\newglossaryentry{o}{name=\ensuremath{o},description={other actor index $o$},sort=so, type=symbols}
\newglossaryentry{j}{name=\ensuremath{j},description={other actor index $j$},sort=sj, type=symbols}
\newglossaryentry{t}{name=\ensuremath{t},description={time step / index $t$},sort=st, type=symbols}
\newglossaryentry{m}{name=\ensuremath{m},description={predicted mode $m$},sort=sm, type=symbols}
\newglossaryentry{n}{name=\ensuremath{n},description={node $n$},sort=sn, type=symbols}
\newglossaryentry{l}{name=\ensuremath{l},description={lane node index $l$},sort=sl, type=symbols}
\newglossaryentry{e}{name=\ensuremath{e},description={edge index $e$},sort=se, type=symbols}
\newglossaryentry{gt}{name=\ensuremath{\text{gt}},description={ground truth},sort=sgt, type=symbols}
\newglossaryentry{fut}{name=\ensuremath{\text{fut}},description={future},sort=sfut, type=symbols}
\newglossaryentry{hist}{name=\ensuremath{\text{hist}},description={history / past},sort=shist, type=symbols}

\newcommand{\ix}{\gls{i}}
\newcommand{\jx}{\gls{j}}

\newcommand{\mx}{\gls{m}}

\newcommand{\nx}{\gls{n}}

\newcommand{\Mx}{\gls{M}}

\newcommand{\traj}{\gls{traj}}

\newglossaryentry{U}{name=\ensuremath{\mathcal{U}\left(x;a,b\right)},description={uniform distribution over the sample space $[a, b]$},sort=xu, type=symbols}
\newglossaryentry{N}{name=\ensuremath{\mathcal{N}\left(x;\mu,\Sigma\right)},text={\ensuremath{\mathcal{N}}},description={(multivariate) normal distribution with parameters $\mu$ and $\Sigma$},sort=xn, type=symbols}
\newglossaryentry{HN}{name=\ensuremath{\mathcal{HN}\left(x;\sigma\right)},text={\ensuremath{\mathcal{HN}}},description={half-normal distribution with parameter $\sigma$},sort=xhn, type=symbols}
\newglossaryentry{Laplace}{name=\ensuremath{\text{Laplace}\left(x;\mu, b\right)},text={\ensuremath{\text{Laplace}}},description={Laplace distribution with parameters $\mu$, $b$},sort=xhn, type=symbols}

\newglossaryentry{disttraj}{name=\ensuremath{\text{dist}\left(\hat{\tau}_{i,m}, \tau_{i,\text{gt}}\right)},text=\ensuremath{\text{dist}},description={Distance measure between trajectories $\hat{\tau}_{i,m}$ and $\tau_{i,\text{gt}}$},sort=fdisttraj, type=symbols}
\newglossaryentry{neigh}{name=\ensuremath{\mathcal{N}(\nx)},text=\ensuremath{\mathcal{N}},description={Neighboring nodes of node \nx},sort=fneigh, type=symbols}
\newglossaryentry{MLPfunc}{name=\ensuremath{\text{MLP}\left(F_\text{in}\right)},text=\ensuremath{\text{MLP}},description={Application of a multi-layer perceptron (MLP) onto input features $F_\text{in}$},sort=fmlp, type=symbols}
\newglossaryentry{TCNfunc}{name=\ensuremath{\text{TCN}\left(F_\text{in}\right)},text=\ensuremath{\text{TCN}},description={Application of a temporal convolutional network (TCN) onto input features $F_\text{in}$},sort=ftcn, type=symbols}
\newglossaryentry{concat}{name=\ensuremath{\text{concat}\left(F_{\text{in},1}, \ldots, F_{\text{in},n}\right)},text=\ensuremath{\text{concat}},description={Concatenates input feature vectors/matrices $F_\text{in},\cdot$},sort=fconcat, type=symbols}
\newglossaryentry{fADE}{name=\ensuremath{\text{ADE}(\tau_a, \tau_b)},text=\ensuremath{\text{ADE}},description={Mean over the euclidean distances between the points in $\tau_a$ and $\tau_b$},sort=fade, type=symbols}
\newglossaryentry{softmax}{name=\ensuremath{\text{softmax}(a)},text=\ensuremath{\text{softmax}},description={Softmax operation on a vector $a$},sort=fsoftmax, type=symbols}

\newglossaryentry{Att}{name=\ensuremath{\text{Att}(Q,K,V)},text=\ensuremath{\text{Att}},description={Attention function on keys $K$, values $V$, and queries $Q$},sort=fatt, type=symbols}
\newglossaryentry{relu}{name=\ensuremath{\text{ReLU}(\cdot)},text=\ensuremath{\text{ReLU}},description={Rectified linear unit. Equivalent to $\max(0, \cdot)$.},sort=frelu, type=symbols}
\newglossaryentry{leakyrelu}{name=\ensuremath{\text{LeakyReLU}(\cdot)},text=\ensuremath{\text{LeakyReLU}},description={Leaky Rectified linear unit. Equivalent to $\max(0, \cdot) + s \min(0, \cdot)$, where $s$ is the negative slope.},sort=fleakyrelu, type=symbols}
\newglossaryentry{sigma}{name=\ensuremath{\sigma(\cdot)},text=\ensuremath{\sigma},description={Placeholder for an activation function, such as ReLU},sort=fsigma, type=symbols}
\newglossaryentry{sigmacomp}{name=\ensuremath{\phi(\cdot)},text=\ensuremath{\phi},description={Sequential application of an activation function, such as ReLU, and layer normalization},sort=fphi, type=symbols}
\newglossaryentry{concatsym}{name=\ensuremath{\normalconc},text=\ensuremath{\normalconc},description={Concatenation of vectors/matrices along an axis},sort=fconcatsym, type=symbols}
\newglossaryentry{min}{name=\ensuremath{\min(a,b)},text=\ensuremath{\min},description={Calculates the minimum of two numbers $a, b$},sort=fmin, type=symbols}
\newglossaryentry{max}{name=\ensuremath{\max(a,b)},text=\ensuremath{\max},description={Calculates the maximum of two numbers $a, b$},sort=fmax, type=symbols}

\glsdisablehyper

\usepackage{graphicx}
\graphicspath{{./img/}{../img/}}

\newlength{\figurewidth}

\newlength{\figureheight}

\newcommand{\thrOne}{\ensuremath{\theta_\text{gt}}}
\newcommand{\thrTwo}{\ensuremath{\theta_\text{type}}}
\newcommand{\thrThree}{\ensuremath{\theta_\text{AOI}}}

\title{\LARGE \bf
Graph-based Trajectory Prediction with Cooperative Information$^\ast$
}

\author{Jan~Strohbeck\orcidlink{0000-0002-2210-1676}, Sebastian Maschke\orcidlink{0009-0005-2770-8871}, Max Mertens\orcidlink{0000-0003-1422-0816}, and Michael~Buchholz\orcidlink{0000-0001-5973-0794}%
	\thanks{$\ast$ This work was financially supported by the Federal Ministry of Economic Affairs and Energy of Germany within the program ``Highly and Fully Automated Driving in Demanding Driving Situations'' (project LUKAS, grant number 19A20004F).}%
	\thanks{The authors are with the Institute of Measurement, Control and Microtechnology,
		Ulm University, D-89081 Ulm, Germany \newline
		{\{first\,\,name.last\,\,name@uni-ulm.de\}}
	}%
}

\begin{document}

\maketitle
\thispagestyle{empty}
\pagestyle{empty}

\begin{abstract}
For automated driving, predicting the future trajectories of other road users in complex traffic situations is a hard problem.
Modern neural networks use the past trajectories of traffic participants as well as map data to gather hints about the possible driver intention and likely maneuvers.
With increasing connectivity between cars and other traffic actors, cooperative information is another source of data that can be used as inputs for trajectory prediction algorithms.
Connected actors might transmit their intended path or even complete planned trajectories to other actors, which simplifies the prediction problem due to the imposed constraints.
In this work, we outline the benefits of using this source of data for trajectory prediction and propose a graph-based neural network architecture that can leverage this additional data.
We show that the network performance increases substantially if cooperative data is present.
Also, our proposed training scheme improves the network's performance even for cases where no cooperative information is available.
We also show that the network can deal with inaccurate cooperative data, which allows it to be used in real automated driving environments.
\end{abstract}

%

\section{Introduction}

Predicting the future behavior and trajectories of other road users is a key challenge for automated driving, especially in urban environments.
Often, neural networks, which have been trained on large datasets of recorded trajectories, are used for this task.
These networks can pick up on subtle cues in the road users' trajectories, which allows for an accurate prediction of their future trajectories.
Alongside the trajectories, data like static map information and traffic signalization data can improve the prediction.

An area that has not received much attention yet is the usage of cooperative information for trajectory prediction.
Currently, there are already some proposed standards for vehicle-to-vehicle communication for cooperative, connected driving, for example the Cooperative Awareness Basic Service \cite{CAM} or the Vulnerable Road Users (VRUs) Basic Service (VBS) \cite{VAM}.
With these services, road users can notify others of their position, and, important for this work, of their intended path or trajectory (in case of VBS).
Additionally, under work item DTR/ITS-00185, the European Telecommunications Standards Institute (ETSI) is currently working on a service for maneuver coordination, which would enable road users to exchange information about their intended paths, and allow for coordination and negotiation of their future trajectories.

The integration of this additional data into trajectory prediction has not yet been addressed in the research.
However, if future trajectories or paths of some traffic participants are already known (with some margin for error), this constrains the future trajectories of the other road users.
Without the additional information, the prediction network would have to take into account more possible evolutions of the given traffic scenario.
Hence, by using the cooperative information, the prediction task can become easier and the network's results can become more accurate.
For this, in this work, we propose an architecture for a neural network as well as a corresponding training scheme, which is the first that can flexibly make use of this additional information.
We evaluate our approach on the large-scale Argoverse Motion Forecasting Dataset \cite{argoverse} and show that the usage of the cooperative information benefits the accuracy of the network's predictions.

\section{Related Work}

Trajectory prediction for automated driving is a topic which has received and continues to receive much research attention. Many approaches use actor historic states and map data as the primary inputs for the prediction \cite{lanegcn,iros1,vectornet}.
However, these works then do not take into account additional information about route options or planned trajectories that might be transmitted by some actors.
Some works, however, condition on an intermediate goal for the actor to improve the prediction.
For example, TNT \cite{tnt} first tries to predict goal points for the actor and then predicts trajectories conditioned on these goal points.
During training, the ground truth trajectory endpoint is used to guide the learning of the conditional motion prediction.
Similarly, GoalNet \cite{goal1} conditions proposals on goals and goal paths that are extracted from map information.
WIMP \cite{whatif} can condition actor prediction on reference polylines using polyline attention.
With this, it can evaluate hypothetical scenarios given a path that an actor should follow.
PGP \cite{pgp} conditions the prediction output on lane traversals which are estimated by a policy network. Additionally, the output can be conditioned on a latent vector that represents a motion profile.

Some works also condition the prediction output on actor trajectories.
For example, M2I \cite{m2i} can condition the prediction of one actor on a future trajectory of another actor, and shows that this can decrease the prediction error significantly.
Similarly, \cite{ego1} train a network that is able to generate conditional predictions based on a query agent's future trajectory, and also show that the additional information leads to an increased performance.
However, only one actor is selected as the query agent per scene, while our approach can handle any number of actors with cooperative information.
RECUP Net \cite{recupnet} employs a recursive structure in which intermediate trajectory predictions by the network are fed into another encoder network, which allows it to better model inter-vehicle interactions.
\cite{mfp} employs recurrent neural networks to iteratively and simultaneously predict all actors in a scene. This can be used to generate predictions which can be conditioned on given trajectories for individual actors. However, the prediction is unimodal, i.e., only a single trajectory is predicted for each actor, and it can not easily handle path information.
CGTP \cite{cgtp} uses a goal prediction network and a trajectory proposal network, where, for two interacting agents, the prediction of one actor can be conditioned on goal points of another actor.
Here, the ground truth trajectories are used during training to train the conditional interaction.
In contrast, our work does not explicitly condition on intermediate outputs or goals, but provides possibly inaccurate paths or future trajectories as additional inputs to the network, which can then decide on how much it wants to trust the cooperative information after fusing it with the historic states and map data.
Also, our network is not reliant on the additional information, but can use it efficiently if it is present.

Some other works condition on possible future actions or trajectories of the ego vehicle.
For example, PiP \cite{pip} first creates candidates of planned trajectories for the ego vehicle, and then conditions the prediction on those plans.
PRECOG \cite{precog} conditions the prediction of other traffic actors on the goals of the ego vehicle, and shows that knowledge of that goal can improve the prediction of the other actors.
In contrast, our work can handle not only explicit trajectories or goals for one actor in the scene, but can rather deal with both path inputs and trajectories, for any number of actors in the scene, not just the ego vehicle.

Approaches that use communicated data for trajectory prediction are rare. \cite{gps1,gps2} use (D)GPS based localization data that is shared between traffic actors, but this is done in an environment perception stage before the trajectory prediction, and no trajectory or path information is shared.

\section{Method}

In this section, we discuss the problem setting, our proposed network architecture, as well as the training scheme.

\subsection{Problem Formulation}

For the problem of trajectory prediction, the goal is to predict future states of actors in the perceived environment.
In the following, it is assumed that the perception of the environment is performed at equally-spaced time intervals.
Then, $\jx = 0$ denotes the current time index, $\jx < 0$ denotes time steps in the past, and $\jx > 0$ denotes time steps in the future.
The actors in a traffic scene can be divided into four different kinds: The ego vehicle, connected automated vehicles, connected actors, and other, non-connected actors.
Connected automated vehicles run a trajectory planning algorithm locally, and transmit their planned trajectories to the ego vehicle, on which the prediction algorithm is run.
Alternatively, actors might only send a path instead of a complete trajectory.
For example, when an actor is only cooperative but not automated, i.e., a human is driving, only the route may be transmitted, which is known from the actor's navigation computer (or similar devices).
For connected automated vehicles, no trajectory prediction has to be performed, since their planned trajectories were already transmitted.
Connected (but not automated) vehicles have to be predicted, but their transmitted path information can be used to improve the prediction.
Thus, we can divide the set of all actors \gls{setall} into two sets, the set of actors that have to be predicted \gls{setpred} and the set of other actors \gls{setother}, whose planned trajectories were already transmitted.
For all actors $\ix \in \gls{setall}$, we assume that historic trajectories \gls{trajihist} with $-\gls{thisti} < \jx \leq 0$ are available.
For this, $\gls{thisti}$ denotes the number of historic states that are available for actor \ix.
Here, a trajectory \traj{} is defined as a set of states \gls{state} at corresponding time steps $\jx$, where states \gls{state} contain the actor's position.
Then, the task is to estimate the ground truth future trajectories \gls{trajigt} of actors $\ix \in \gls{setpred}$.
A prediction algorithm can estimate \gls{trajigt} by predicting a set of \Mx{} trajectories \gls{trajpredm}, where $\mx \in \{ 1, \ldots \Mx \}$, alongside probabilities \gls{pmodei} for each predicted trajectory.
For the prediction, historic trajectories \gls{trajihist}, future trajectories \gls{trajifut} of connected automated actors, as well as paths \gls{pathifut} of connected actors, can be available.
Here, a path contains a sequence of positions but no time information.
Static map information \gls{map}, which contains information about lanes and their geometry, can be used as an additional input.

\subsection{Training Scheme}

\begin{figure*}[tb]
\vspace{2mm}%
\begin{minipage}{0.46\columnwidth}%
\begin{subfigure}[b]{\linewidth}
\centering
\includegraphics[width=\linewidth]{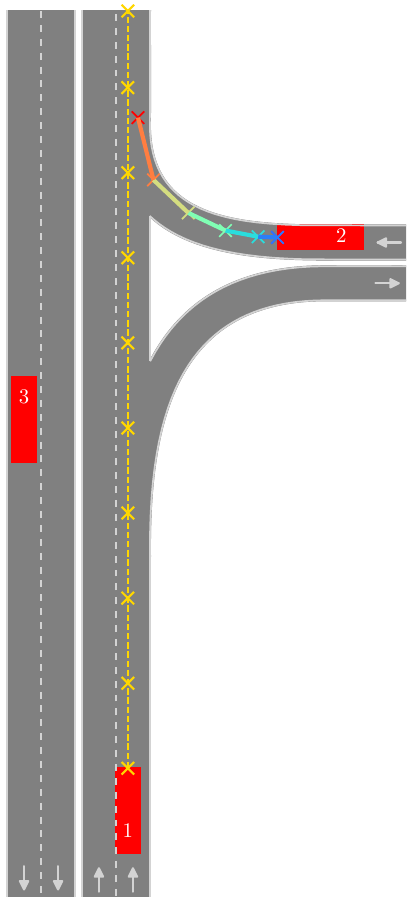}
\caption{Example scene.}
\label{fig:example_scene}
\end{subfigure}%
\end{minipage}%
\hfill
\begin{minipage}{0.46\columnwidth}%
\begin{subfigure}[b]{\linewidth}
\centering
\includegraphics[width=0.98\linewidth]{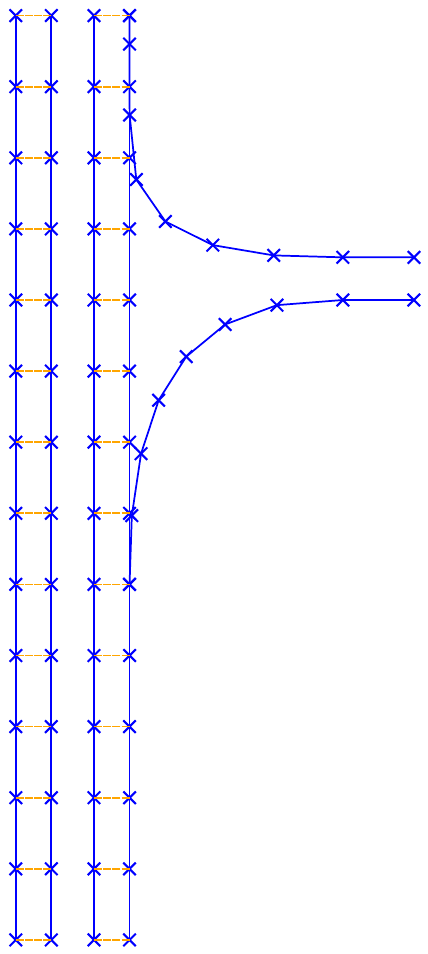}
\caption{Lane graph \gls{graphl}.}
\label{fig:lanes_graph}
\end{subfigure}%
\end{minipage}%
\hfill
\begin{minipage}{\columnwidth}%
\begin{subfigure}[b]{0.3\linewidth}
\centering
\includegraphics[width=0.866666\linewidth]{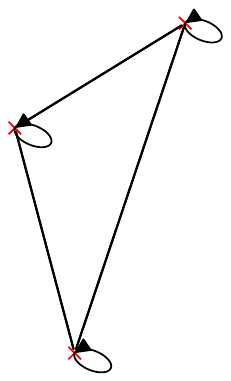}
\caption{Actor graph \gls{graphact}.}
\label{fig:a2a}
\end{subfigure}%
\hfill%
\begin{subfigure}[b]{0.2\linewidth}
\centering
\includegraphics[width=0.6\linewidth]{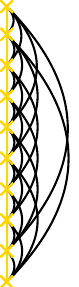}
\caption{Path graph \gls{graphpath}.}
\label{fig:p2p}
\end{subfigure}%
\hfill%
\begin{subfigure}[b]{0.3\linewidth}
\centering
\includegraphics[width=\linewidth]{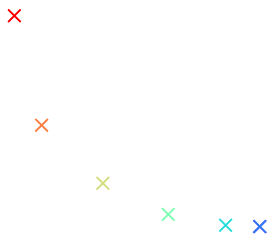}
\caption{Trajectory graph \gls{graphtraj}.}
\label{fig:trajs}
\end{subfigure}%
\newline
\begin{subfigure}[b]{0.50\linewidth}
\includegraphics[width=0.8\linewidth]{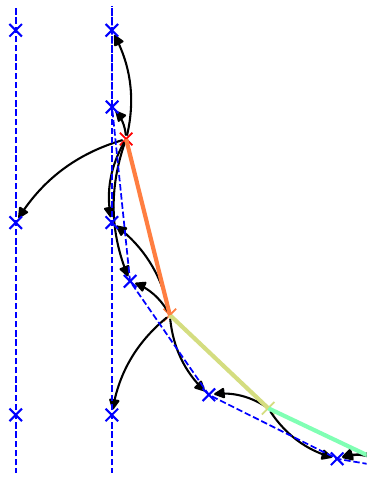}
\caption{Trajectory-to-lane (T2L).}
\label{fig:f2l}
\end{subfigure}%
\hfill%
\begin{subfigure}[b]{0.5\linewidth}
\includegraphics[width=\linewidth]{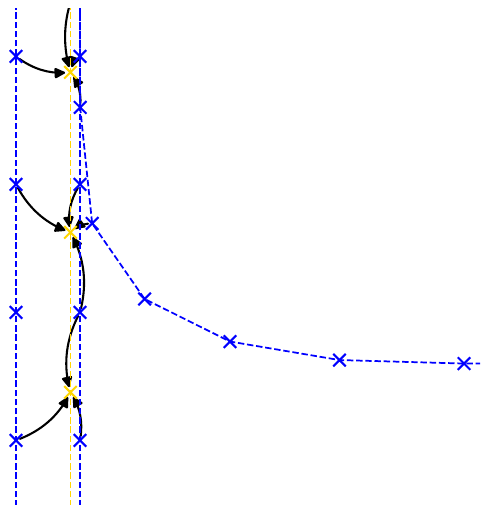}
\caption{Lane-to-path (L2P).}
\label{fig:l2p}
\end{subfigure}%
\end{minipage}%
\caption{(a) shows an example scene with cooperative information. Actor 1 transmits a path (gold line) along which it wants to move. Actor 2 transmits a complete trajectory (rainbow colors indicate time information), which indicates that it will accelerate and merge onto Actor 1's lane. When predicting Actor 1, the prediction problem is constrained by its path (no hypothesis for turning right is necessary) and the movement of Actor 2 (velocity has to be adjusted). Actor 3 transmits no cooperative information and has to be predicted conventionally. For the corresponding lane graph in (b), lane nodes and predecessor/successor connections are in blue, left/right connections are dashed orange. Self- and dilated connections are omitted for clarity. (c), (d), (e) show the corresponding actor graph, path graph and trajectory graph, respectively. (f) and (g) show two of the novel fusion modules. The T2L module (f) updates lane nodes with data from nearby trajectory graph nodes, whereas L2P (g) updates path nodes with data from nearby lane nodes.}
\label{fig:example_scene_full}
\end{figure*}

For training, we use a supervised approach, in which we learn on a dataset of recorded vehicle trajectories.
However, there are currently no datasets with connected vehicles that transmitted their planned trajectories.
To solve this problem, we emulate the planned trajectories by using some of the ground truth trajectories of actors as inputs to the network.
For example, if a scene contains 5 actors, we might try to predict the movement of actors 1 and 3 normally, and use the trajectories of actor 2, 4 and 5 as cooperative inputs ($\tau_{2,\text{fut}}, \tau_{4,\text{fut}}, \tau_{5,\text{fut}}$).

We denote the fraction of actors whose trajectories are used as $\thrOne \in [0, 1]$, and use a uniform distribution to sample it randomly for each training sample.
For example, $\thrOne = 0.2$ would correspond to treating \SI{20}{\percent} of actors as cooperative and using their trajectories.
This random sampling also acts as a data augmentation which is beneficial to the training.

Since the network should handle both the case that an actor transmits a complete trajectory \gls{trajifut} as well as the case that it only transmits an intended path \gls{pathifut}, we introduce a second parameter $\thrTwo \in [0, 1]$.
This fraction is also sampled randomly for each training sample, and divides the set of cooperative actors into $N_\text{traj} = \lfloor N \thrTwo + \frac{1}{2} \rfloor$ actors with trajectories and $N - N_\text{traj}$ actors with only paths, where $N$ is the number of cooperative actors sampled using \thrOne.
Actors with cooperative paths are still predicted, while actors with cooperative trajectories are, of course, not predicted.

Using the exact ground truth trajectories would lead to the network relying heavily on the accuracy of the cooperative information.
In reality, however, planned trajectories are not exactly equal to the actually driven trajectories.
For example, a vehicle might have to slow down due to unforeseen changes in the traffic scenario, such as new actors entering the scene, or it might be able to drive faster because the original planned trajectory was planned under conservative assumptions.
To mirror this, we do not use the exact ground truth trajectories, but use them only in an augmented way.

Our proposed augmentation is to speed up or slow down the trajectory by a random factor $\beta \in [0, 2]$.
Slowing down is performed by linear interpolation of the ground truth trajectory, while speeding up requires inter- and extrapolation.
This assumes that the original path that the actor wants to follow is fixed and unchanging.
This is a realistic assumption, since, in most cases, a route is selected at the start of an actor's journey, and route changes are rare compared to simple speed adjustments.

Paths are constructed from augmented trajectories by sampling points at \SI{2}{\meter} intervals.
Hence, for $\beta < 1$ or $\beta > 1$, the path can become shorter or longer, respectively, which prevents the network from just predicting until the end of the path, which would be a kind of overfitting.

\subsection{Network Architecture}
\label{sec:arch}

\begin{figure*}[tb]
	\vspace{2mm}
  \centering
  \includegraphics[width=0.8\textwidth]{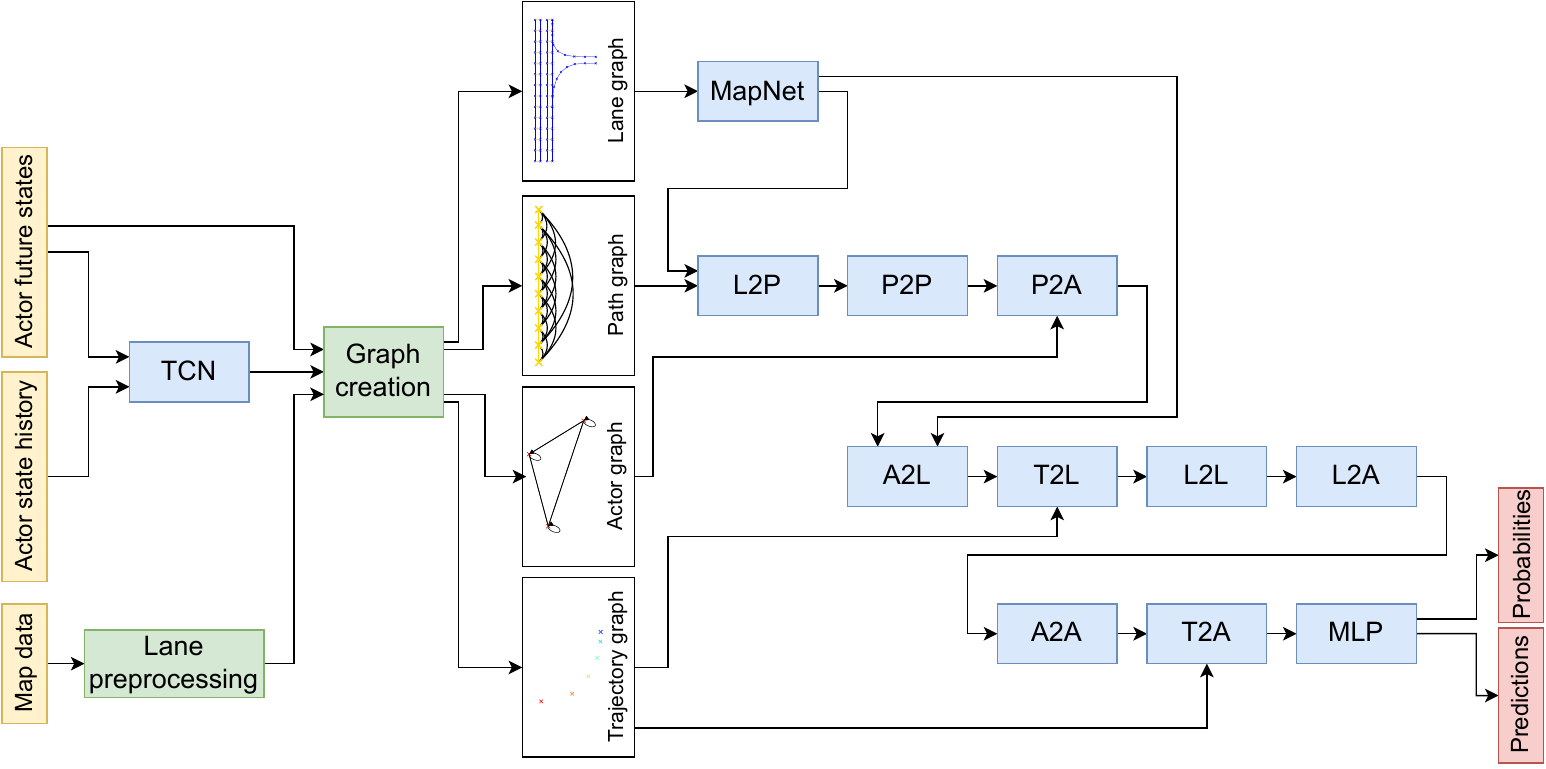}



  \caption{Overview of the network architecture. Input data is colored in light orange, trainable network layers are colored blue, network outputs are red, and other processing modules are in green. The network is composed of a temporal convolutional network (TCN) and a graph creation step, after which a series of graph convolution modules is applied, before a multi-layer perceptron (MLP) calculates the final output from the resulting actor node features.}
  \label{fig:model_architecture}
\end{figure*}

In this work, we model the input data as heterogenous graphs.
For this, we construct four individual graphs, namely actor graph \gls{graphact}, lane graph \gls{graphl}, trajectory graph \gls{graphtraj}, and path graph \gls{graphpath}.
This is visualized in Fig.~\ref{fig:example_scene_full}, where an example traffic scene with cooperative information is visualized.
In the scene, one actor transmits its path, one transmits a complete trajectory, and one is not cooperative.
First, map information is used to construct \gls{graphl} (see Fig.~\ref{fig:lanes_graph}), which consists of lane centerline nodes that are connected based on successor/predecessor/left/right relations, exactly as in \cite{lanegcn}.
In the actor graph (Fig.~\ref{fig:a2a}), nodes are connected to other actor nodes if they are less than a threshold euclidean distance \gls{threshconnact} apart, similar to \cite{lanegcn}.
In the novel path graph \gls{graphpath}, nodes correspond to path points of cooperative paths and are connected based on successor/predecessor relationships (see Fig.~\ref{fig:p2p}), similar to the lane graph.
We adopt the scheme of \cite{lanegcn}, where lane graph nodes are connected also via dilated connections which skip a certain number of nodes, and use it for the path graph \gls{graphpath} as well.
As can be seen in Fig.~\ref{fig:p2p}, nodes in \gls{graphpath} are connected by $n$-dilated connections if they are farther apart but reachable by traveling along a predecessor/successor type edge $n$ times.
We use dilation sizes $[1, 2, 4, 8]$ for \gls{graphpath} and $[1, 2, 4, 8, 16, 32]$ for \gls{graphl}.

The trajectory graph \gls{graphtraj} contains nodes which correspond to future trajectory points of cooperative actors (see Fig.~\ref{fig:trajs}).
That is, if an actor transmits a sequence of trajectory points, those points correspond to nodes in \gls{graphtraj}.
In the trajectory graph, nodes are not connected.
Rather, those nodes are only connected to nodes in the other two graphs.
For this interconnection between graphs, we adopt a fusion scheme similar to LaneGCN \cite{lanegcn}.
That is, we introduce additional edges between the graphs based on proximity (like for the actor graph connections), and use a graph convolutional network (GCN) that uses message passing to exchange information between the graphs.
For this, we adopt the concept of the fusion networks lanes-to-lanes (L2L), actors-to-lanes (A2L), lanes-to-actors (L2A), and actors-to-actors (A2A) from \cite{lanegcn}.
To integrate information from the new graphs \gls{graphtraj} and \gls{graphpath}, we introduce the new fusion networks trajectories-to-lanes (T2L), trajectories-to-actors (T2A), lanes-to-paths (L2P), paths-to-paths (P2P), and paths-to-actors (P2A).

The full processing sequence, which consists of a temporal convolutional network (TCN) \cite{lea2016temporal}, a graph creation step, and subsequent GCN layers, is shown in Fig.~\ref{fig:model_architecture}.
In the following, we outline the data flow through the different modules.
The graph convolutions used in the GCN layers are explained in Section~\ref{sec:convs}.
First, the actor states (history and future) are combined to a single trajectory, where the future states are taken from cooperative trajectories.
The resulting trajectory is padded with zeros where no data is available or in case the actor has no cooperative future trajectory.
For actors with only a future path instead of a trajectory, the future states are zeros as well, since no time information is available.
The combined trajectory is then differentiated, so that it contains the differences between trajectory states.
Also, a column with binary values is added, which indicate if a state is available or not (similar to \cite{lanegcn}).
The combined trajectory is then passed through a TCN, which extracts temporal features.
The features at the current time index are used by the graph creation module as the feature vectors for the nodes in \gls{graphact}.
Similarly, the features corresponding to future time indices are used as the features of the nodes in \gls{graphtraj}.
Additionally, the graph creation module creates the other graphs \gls{graphl} and \gls{graphpath}.
Note that lane preprocessing is done only once and, thus, does not incur much latency.

As the first graph operation, a MapNet similar to \cite{lanegcn} is used to process the lane graph by updating the features of lane graph nodes based on neighboring nodes.
Then, lane nodes are used to update features of nodes in the path graph with the L2P module (see Fig.~\ref{fig:l2p}).
The P2P module is then used to process the data contained in each path by propagating along the path using successor/predecessor and dilated connections (see Fig.~\ref{fig:p2p}).
Finally, the P2A module transmits data from each actor's future path (if it has one) to the corresponding actor.
For this, path points in \gls{graphpath} are connected to the corresponding actor node in \gls{graphact}.
Then, A2L is used to pass information from \gls{graphact} to \gls{graphl}, so that the lane nodes can receive information about nearby actors.
With this, the network can know if actors currently occupy certain lane segments, and what their intentions might be.
L2L then, like MapNet, operates on the lane graph only, and is used to propagate information along the lanes.
With T2L, the network can additionally propagate the information about future positions of actors to the lane nodes, helping it estimate the future occupancy of lane nodes.
L2A then passes information from the lane nodes back to the actors.
After the actor features have been updated in this way, the remaining two modules are A2A and T2A.
A2A operates only on the actor graph and is used to model interactions between actors (as in \cite{lanegcn}).
The final module, T2A, similarly, allows the network to directly use features from future actor states for the prediction of interactions between actors.

The novel modules, namely T2L, T2A, L2P, P2P, and P2A, thus, enable the network to use the additional information from the trajectory graph and path graph, thereby improving its prediction performance.
The final prediction of actor trajectory hypotheses \gls{trajpredm} and mode probabilities \gls{pmodei} is done by a simple multi-layer perceptron (MLP) with two linear layers and a ReLU activation in between.

\subsection{Graph Convolutions}
\label{sec:convs}

For MapNet, L2L and P2P, we use the multi-scale LaneConv operator from \cite{lanegcn}
\begin{equation}
	\mathbf{x}_{\ix}' = \sum_{c \in \gls{conns}} \sum_{\jx\in \gls{neigh}_c(\ix)} \mathbf{x}_{\jx} \mathbf{W}_c,
\end{equation}
where $\mathbf{x}_{\ix}$ are the node features of graph node \ix.
\gls{conns} means the set of possible connection types, i.e., successor, predecessor, and except for P2P, left and right.
Additionally, \gls{conns} contains connection types that correspond to $n$-dilated connections (see sec.~\ref{sec:arch}).
This is analogous to the dilated LaneConv operator from \cite{lanegcn}.
$\mathscr{N}_c(\ix)$ is then the set of indices of neighboring nodes of \ix{} by type $c$, and $\mathbf{W}_c$ are learnable weight matrices for each $c$.
Like \cite{lanegcn}, for each of the three modules, we stack 4 LaneConv layers with residual connections.

For the fusion modules L2A, A2A, A2L, L2P, P2A, T2A and T2L, we update the node features with GATv2 \cite{gatv2} with 8 attention heads.
As edge features, we use $[ \Delta x, \Delta y, \sqrt{(\Delta x)^2 + (\Delta y)^2} ]$, which corresponds to the difference of x/y position of the nodes, and the euclidean distance.
For the novel modules T2L and T2A, we additionally use $t_j/H$ as a feature, which is the fraction of the time index $t_j$ of a node \jx{} in \gls{graphtraj} and the total number of predicted steps $H$.
This allows the network to assess which time step the data belongs to.
For each fusion module, we stack two GCN layers with residual connections, as in \cite{lanegcn}.
For the connection thresholds \gls{threshconnact}, we use \SI{7}{\meter} for L2A, A2L, L2P, P2A and T2L, \SI{6}{\meter} for L2L, and \SI{100}{\meter} for A2A and T2A.

\subsection{Loss Function}

For the loss function, we use
\begin{equation}
	\mathcal{L} = \mathcal{L}_\text{pos} + \mathcal{L}_\text{class},
\end{equation}
which is a linear combination of position loss $\mathcal{L}_\text{pos}$ and mode classification loss $\mathcal{L}_\text{class}$.
We use a smooth $\text{L}_1$ loss \cite{lanegcn} for the position and a cross-entropy loss for the mode probabilities.

\section{Evaluation}

In this section, we describe the experimental setup, the evaluation metrics and discuss our results.

\subsection{Experimental Setup}

For training and evaluation, we use the Argoverse Motion Forecasting Dataset \cite{argoverse} in version 1.1.
The dataset contains 205,943 traffic scenes for training and 39,472 scenes for validation.
Each scene contains a certain number of actors, whose historic trajectories of the last two seconds are given.
For each scene, one actor is the \emph{actor of interest} (AOI), whose trajectory for the next three seconds has to be predicted.
While the other actors in the scene are also predicted during training, evaluation metrics are calculated and reported only for the AOI to ensure comparability with other works.
Note that during training, actors whose ground truth is used as a cooperative trajectory input can never be part of the set of actors whose trajectories are to be predicted, as this would lead to overfitting.
However, actors whose ground truth is used to construct a cooperative path are to be predicted, since the exact trajectory is not known.
For the evaluation, we exclude the AOI from the set of actors that can be made cooperative using \thrOne, but instead use a separate parameter \thrThree{} to denote evaluations where the AOI is made cooperative ($\thrThree = \checkmark$) or not ($\thrThree = \times$).
This allows to distinguish between cases where knowledge gained from other cooperative actors benefits the prediction of the (non-cooperative) AOI, from cases where the AOI is itself cooperative (only in the form of a transmitted path), which also helps to predict the AOI's trajectory.

We use $M=6$, i.e., the network predicts 6 trajectory hypotheses, along with 6 probabilities.
Each graph node is represented by 256 features.
For \gls{graphpath} and \gls{graphl}, this is achieved using a MLP to encode the input features (similar to \cite{lanegcn}), while for \gls{graphact} and \gls{graphtraj} this is done by using the features extracted by the TCN.
For the TCN, we use three convolutional layers with a kernel size of 3.
To have a smaller number of graph nodes in \gls{graphtraj}, we use only every third point, which does not decrease the performance noticeably.
The MLP that calculates \gls{trajpredm} and \gls{pmodei} consists of two layers, where the first layer calculates 4096 hidden features.
The network was implemented using PyG \cite{pyg} and trained for 150,000 iterations with a batch size of 64, on an NVIDIA RTX 3090 GPU.
For the optimizer, we use Adam with a 1cycle learning rate schedule \cite{onecycle} and maximum learning rate of 1e-3.

\subsection{Evaluation Metrics}

When comparing the set of $M$ predicted trajectories to the ground truth trajectory, one can determine the predicted trajectory that fits best to the ground truth.
As evaluation metrics, we can then evaluate the minimum average displacement error (minADE), minimum final displacement error (minFDE), and the miss rate (MR) \cite{argoverse}.
minADE is the average euclidean distance between that best trajectory and the ground truth trajectory.
minFDE measures the euclidean distance between the end point of the best trajectory and the end point of the ground truth trajectory.
Both metrics are measured in meters.
The miss rate is the percentage of scenarios, in which no trajectory hypothesis with a minFDE of \SI{2}{\metre} or less exists.
Additionally, one can calculate these metrics for different subsets of the predicted trajectories, i.e., taking only the $K$ most probable trajectories (by the network's estimation) \cite{argoverse}.

\subsection{Results}

\begin{table*}[tb]
\vspace{0.12cm}
\centering
\caption{\uppercase{Experimental results on the Argoverse Motion Forecasting validation set.}}
\label{tab:results}
\begin{tabular}{lc|rrc|ccc|ccc}
 & & \multicolumn{3}{c|}{Parameter values} & \multicolumn{3}{c|}{$K=6$} & \multicolumn{3}{c}{$K=1$} \\
& Config & \thrOne & \thrTwo & \thrThree & minADE & minFDE & MR & minADE & minFDE & MR \\
\hline
LaneGCN \cite{lanegcn} & & N/A & N/A & N/A & 0.71 & 1.08 &  & 1.35 & 2.97 &  \\
TNT \cite{tnt} & & N/A & N/A & N/A & 0.73 & 1.29 &  &  &  &  \\
Baseline & & N/A & N/A & N/A & 0.734 & 1.137 & \SI{11.3}{\percent} & 1.446 & 3.242 & \SI{53.0}{\percent} \\
\hline
Ours & I & \SI{0}{\percent} & \SI{0}{\percent} & $\times$ & 0.715 & 1.093 & \SI{10.4}{\percent} & 1.390 & 3.092 & \SI{51.2}{\percent} \\
Ours & II & \SI{100}{\percent} & \SI{0}{\percent} & $\times$ & 0.711 & 1.078 & \SI{10.3}{\percent} & 1.390 & 3.096 & \SI{51.3}{\percent} \\
Ours & III & \SI{100}{\percent} & \SI{100}{\percent} & $\times$ & 0.684 & 1.036 & \SI{9.9}{\percent} & 1.318 & 2.939 & \SI{50.1}{\percent} \\
Ours & IV & \SI{0}{\percent} & \SI{0}{\percent} & $\checkmark$ & 0.584 & 0.753 & \SI{4.1}{\percent} & 1.128 & 2.453 & \SI{45.4}{\percent} \\
Ours & V & \SI{100}{\percent} & \SI{0}{\percent} & $\checkmark$ & 0.580 & 0.742 & \SI{4.0}{\percent} & 1.126 & 2.454 & \SI{45.5}{\percent} \\
Ours & VI & \SI{100}{\percent} & \SI{100}{\percent} & $\checkmark$ & \textbf{0.559} & \textbf{0.715} & \textbf{\SI{3.8}{\percent}} & \textbf{1.080} & \textbf{2.355} & \textbf{\SI{44.5}{\percent}} \\

\end{tabular}
\end{table*}

We evaluate our trained model for different fixed values of \thrOne{} and \thrTwo{}, with $\beta=1$ (no augmentation).
With this, one can see how the network performs with different fractions of cooperative actors per scene (\thrOne), and how the different types of input (\thrTwo) benefit the prediction.
Since the model was trained with random values of \thrOne{} and \thrTwo{}, it should be able to handle any possible combination.
For simplicity, we refer to these different evaluation parameter settings using roman numerals, as shown in the table.
As the baseline, we use a variant of our network without our novel modules, and without using ground truth data as any form of input.
Lower values are better in all cases, and the best results in each metric are formatted in bold.
We can see that the baseline, i.e., no cooperative information, performs worse than configuration I, in which no cooperative information is used.
This means that even without cooperative information, our network receives benefits from our training strategy, particularly from the data augmentation.
Note that our baseline performs slightly worse than LaneGCN and TNT, which use more advanced prediction heads instead of a simple MLP like in our network.
With our training scheme, however, our network is able to overcome that difference without changes to the prediction head (which could still be performed to further improve results).

Our Configurations II and III correspond to the extreme cases of $\thrOne = \SI{100}{\percent}$ of other actors being cooperative and transmitting their paths (II) or their trajectories (III), respectively.
Here, we can see that the additional information is useful when predicting the trajectory for the AOI, as the prediction errors decrease.
However, cooperative trajectories are more useful than cooperative paths, which is reasonable, since they contain concrete timing data.
In configurations IV, V and VI, the AOI is cooperative in the form of a cooperative path.
One can see that the cooperative path information is hugely helpful in predicting the AOI's future trajectory.
Adding cooperative paths (V) or cooperative trajectories (VI) of the other actors in the scene further improves results, with configuration VI achieving the best overall results.
This is of course to be expected, since it receives the most amount of information as the input.
These results indicate that the network is able to flexibly use the cooperative information if it is present, but does not degrade if it is not present.

\begin{figure}[tb]
  \centering
  {\footnotesize%
    \setlength{\figurewidth}{1.00\linewidth}%
    \setlength{\figureheight}{0.4\linewidth}%
\begin{tikzpicture}

\definecolor{darkgray176}{RGB}{176,176,176}
\definecolor{darkorange25512714}{RGB}{255,127,14}
\definecolor{lightgray204}{RGB}{204,204,204}
\definecolor{steelblue31119180}{RGB}{31,119,180}

\begin{axis}[
height=\figureheight,
legend cell align={left},
legend style={
  fill opacity=0.8,
  draw opacity=1,
  text opacity=1,
  at={(0.03,0.03)},
  anchor=south west,
  draw=lightgray204
},
tick align=outside,
tick pos=left,
width=\figurewidth,
x grid style={darkgray176},
xlabel={\(\displaystyle \theta_{\mathrm{gt}}\)},
xmin=-5, xmax=105,
xtick style={color=black},
y grid style={darkgray176},
ylabel shift={0pt},
ylabel={minFDE @ K=6},
ymin=1.03315, ymax=1.09585,
ytick style={color=black}
]
\addplot [semithick, steelblue31119180]
table {%
0 1.093
10 1.09
20 1.088
30 1.087
40 1.087
50 1.084
60 1.084
70 1.081
80 1.079
90 1.079
100 1.078
};
\addlegendentry{$\theta_{\mathrm{type}} = 0\%$}
\addplot [semithick, darkorange25512714]
table {%
0 1.093
10 1.077
20 1.063
30 1.056
40 1.051
50 1.044
60 1.046
70 1.042
80 1.039
90 1.037
100 1.036
};
\addlegendentry{$\theta_{\mathrm{type}} = 100\%$}
\end{axis}

\end{tikzpicture}%
  }%
  \caption{Error when predicting the AOI's trajectory when increasing the amount of cooperative data of other actors (\thrOne) when $\beta=1$ and $\thrThree = \times$. $\thrTwo{} = \SI{0}{\percent}$ (blue) means that cooperative data stems from transmitted path information, while $\thrTwo = \SI{100}{\percent}$ means that it stems from transmitted planned trajectories.}
  \label{fig:sweeps}
\end{figure}
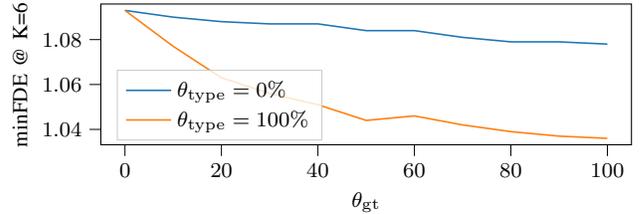

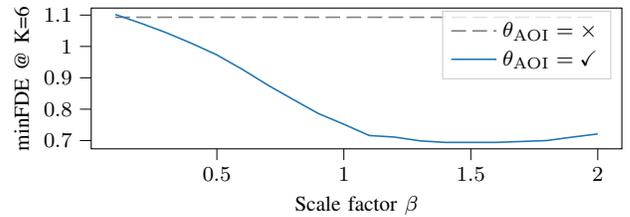
\begin{figure}[tb]
  \centering
  {\footnotesize%
    \setlength{\figurewidth}{1.00\linewidth}%
    \setlength{\figureheight}{0.4\linewidth}%
\begin{tikzpicture}

\definecolor{darkgray176}{RGB}{176,176,176}
\definecolor{gray}{RGB}{128,128,128}
\definecolor{lightgray204}{RGB}{204,204,204}
\definecolor{steelblue31119180}{RGB}{31,119,180}

\begin{axis}[
height=\figureheight,
legend cell align={left},
legend style={fill opacity=0.8, draw opacity=1, text opacity=1, draw=lightgray204},
tick align=outside,
tick pos=left,
width=\figurewidth,
x grid style={darkgray176},
xlabel={Scale factor \(\displaystyle \beta\)},
xmin=0.005, xmax=2.095,
xtick style={color=black},
y grid style={darkgray176},
ylabel shift={0pt},
ylabel={minFDE @ K=6},
ymin=0.6736, ymax=1.1224,
ytick style={color=black}
]
\addplot [semithick, gray, dash pattern=on 5.55pt off 2.4pt]
table {%
0.1 1.093
2 1.093
};
\addlegendentry{$\theta_{\mathrm{AOI}} = \times$}
\addplot [semithick, steelblue31119180]
table {%
0.1 1.102
0.2 1.074
0.3 1.044
0.4 1.01
0.5 0.973
0.6 0.927
0.7 0.877
0.8 0.831
0.9 0.786
1 0.752
1.1 0.716
1.2 0.711
1.3 0.699
1.4 0.694
1.5 0.694
1.6 0.694
1.7 0.697
1.8 0.7
1.9 0.711
2 0.721
};
\addlegendentry{$\theta_{\mathrm{AOI}} = \checkmark$}
\end{axis}

\end{tikzpicture}%
  }%
  \caption{Prediction error for the AOI at $\thrOne = \SI{0}{\percent}$ and $\thrThree = \checkmark$ at varying values of $\beta$. Prediction error at $\thrThree = \times$ is shown as the reference.}
  \label{fig:sweeps_scale}
\end{figure}

This is confirmed by our second evaluation, the results of which are shown in Fig.~\ref{fig:sweeps}.
Here, one can see that when increasing the amount of cooperative paths or cooperative trajectories of actors other than the AOI, the prediction error almost decreases in a linear fashion.
Again, cooperative trajectories are more beneficial than cooperative paths.

To show that the network can handle inaccurate cooperative data, in Fig.~\ref{fig:sweeps_scale} we evaluate the network for different values of $\beta$ when predicting the AOI with an imperfect cooperative path.
We can see that as $\beta$ increases from 0.1 to 1, i.e., the cooperative paths become longer, the network predictions become more accurate.
At $\beta > 1$, the path becomes longer due to extrapolation, and we see that a slight extrapolation still benefits the network.
Increasingly large values $\beta > 1.7$ perform slightly worse due to the simple linear extrapolation during training, which often does not correspond to a feasible path along the lanes.
In real applications with feasible paths, this slight performance drop should not be experienced.
Also, the network still benefits hugely from the path information (compared to the $\thrThree=\times$ reference), even if the extrapolated path is, in sum, infeasible.
The network also benefits from increasingly longer paths, rather than overfitting on the correct path length ($\beta = 1$).

\begin{figure*}[tb]
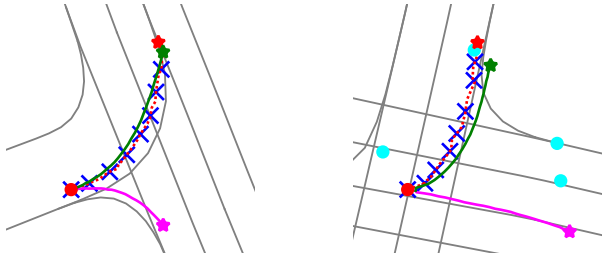
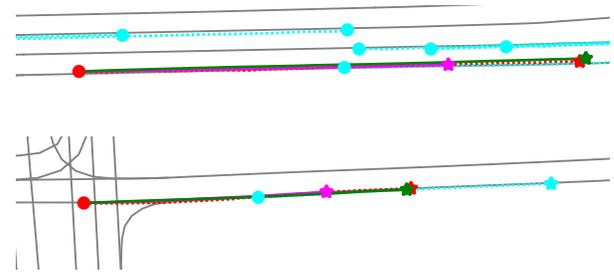

  \vspace{0.2cm}
\begin{subfigure}[b]{0.48\linewidth}
\begin{minipage}{0.46\linewidth}%
\begin{center}
\input{img/examples_18837.pgf}%
\end{center}
\end{minipage}
\hfill
\begin{minipage}{0.46\linewidth}%
\begin{center}
\input{img/examples_37519.pgf}%
\end{center}
\end{minipage}
\caption{With the AOI's path as cooperative information (blue crosses), the network's best prediction (green) follows the path and is close to the ground truth (red).}
\label{fig:example_path}
\end{subfigure}%
\hfill%
\begin{subfigure}[b]{0.48\linewidth}
\begin{center}
\vbox{\input{img/examples_9613.pgf}}%
\vbox{\input{img/examples_19823.pgf}}%
\end{center}
\caption{With the future trajectories of other actors as cooperative information (cyan lines), the network's best prediction (green) is less conservative and closer to the ground truth (red).}
\label{fig:example_traj}
\end{subfigure}%
\caption{Examples where cooperative information improves the prediction of the AOI's trajectory. Circles denote positions of actors, where the AOI is red and other actors are cyan. The AOI's ground truth trajectory is shown as a red dashed line. Best predicted hypotheses with and without cooperative information are in green and magenta, respectively. Trajectory end points are denoted with a star in the corresponding color. Lane centerlines are in grey.}
\label{fig:examples}
\end{figure*}

In Fig.~\ref{fig:examples}, we show examples where the cooperative information proves to be beneficial.
Fig.~\ref{fig:example_path} shows that without the cooperative path information, the network might miss turning maneuvers completely in some cases. With the additional path information, the network correctly predicts the turn, and the prediction error is significantly reduced.
In Fig.~\ref{fig:example_traj}, with the cooperative trajectories of other actors, the prediction of the AOI is improved. This is often because the future speeds of other actors (especially leading cars) are known, which allows for less conservative and more optimized trajectory predictions with lower resulting error.

\section{Conclusions}

In this work, we outlined the problem of predicting trajectories when cooperative information is present.
We showed the benefits of using this data and proposed a training scheme to allow a supervised training for a neural network-based approach.
We also proposed a network architecture which can make use of this additional information, and showed that it is flexibly able to properly use any amount of cooperative data that is provided.
The network can deal with cooperative trajectories as well as cooperative paths, which allows it to handle both fully automated actors, as well as connected actors with human drivers.
Also, the network's performance is improved even for cases where no cooperative information is present, due to the data augmentation included in our training scheme.
In our future work, we want to improve the network architecture to make even better use of the additional information, and to cover more types of input, such as traffic lights.


\bibliographystyle{IEEEtran}
\bibliography{bibliog}

\end{document}